\documentclass[runningheads]{llncs}
\usepackage{graphicx}
\usepackage{siunitx}
\usepackage{hyperref}
\usepackage{color}
\usepackage{amssymb}
\usepackage{algorithm}
\usepackage{algpseudocode}
\usepackage{makecell}
\usepackage{multirow}
\usepackage{lipsum}  
\usepackage{tikz-cd}
\newcommand{\etal}{\textit{et al.}}
\usepackage{ amssymb }

\newcommand{\bfsection}[1]{\vspace*{0.1cm}\noindent\textbf{#1:}}

\definecolor{MyBlue}{rgb}{0,0.08,0.5}
\definecolor{MyRed}{rgb}{0.7,0.02,0.02}
\definecolor{MyOrange}{rgb}{1,0.5,0}
\definecolor{MyPurple}{rgb}{0.6,0.25,0.8}
\definecolor{MyGreen}{rgb}{0.1,0.8,0.1}
\definecolor{mygray}{gray}{0.4}

\usepackage{amsmath}
\usepackage{bbm}
\usepackage{subfigure}
\DeclareMathOperator*{\argmax}{arg max} 

    \makeatletter
\def\@fnsymbol#1{\ensuremath{\ifcase#1\or \dagger\or \ddagger\or
   \mathsection\or \mathparagraph\or \|\or **\or \dagger\dagger
   \or \ddagger\ddagger \else\@ctrerr\fi}}
    \makeatother

\begin{document}

\title{Developmental Stage Classification of Embryos Using Two-Stream Neural Network with Linear-Chain Conditional Random Field}



\author{Stanislav Lukyanenko\inst{1}\thanks{Works were done during the internship at Harvard University.} \and Won-Dong Jang\inst{2} \and Donglai Wei\inst{2} \and Robbert Struyven\inst{2, 5} \and \\ Yoon Kim\inst{6} \and Brian Leahy\inst{2,3} \and Helen Yang\inst{3, 4} \and Alexander Rush\inst{7} \and \\ Dalit Ben-Yosef\inst{9,10} \and Daniel Needleman\inst{2,3,8} \and Hanspeter Pfister\inst{2}}



\newcommand*{\affaddr}[1]{#1} 
\newcommand*{\affmark}[1][*]{\textsuperscript{#1}}
\institute{
\affaddr{\affmark[1] Department of Informatics, Technical University of Munich, Germany} \\ \
\affaddr{\{\affmark[2] School of Engineering and Applied Sciences,} \
\affaddr{\affmark[3] Department of Molecular and Cellular Biology,} \
\affaddr{\affmark[4] Graduate Program in Biophysics\}, Harvard University, USA} \\ \
\affaddr{\affmark[5] University College London, UK} \
\affaddr{\affmark[6] MIT, USA} \
\affaddr{\affmark[7] Cornell University, USA} \\ \ 
\affaddr{\affmark[8] Center for Computational Biology, Flatiron Institute, USA} \\ \
\affaddr{\affmark[9] Lis Maternity Hospital, Tel-Aviv Sourasky Medical Center, Israel} \\ \
\affaddr{\affmark[10] Cell and Developmental Biology, Tel-Aviv University, Israel}\\
\email{stanislav.lukyanenko@tum.de}
}

%
\maketitle   

\begin{abstract}
The developmental process of embryos follows a monotonic order. An embryo can progressively cleave from one cell to multiple cells and finally transform to morula and blastocyst. For time-lapse videos of embryos, most existing developmental stage classification methods conduct per-frame predictions using an image frame at each time step. However, classification using only images suffers from overlapping between cells and imbalance between stages. Temporal information can be valuable in addressing this problem by capturing movements between neighboring frames. In this work, we propose a two-stream model for developmental stage classification. Unlike previous methods, our two-stream model accepts both \textit{temporal} and \textit{image} information. We develop a linear-chain conditional random field (CRF) on top of neural network features extracted from the temporal and image streams to make use of both modalities. The linear-chain CRF formulation enables tractable training of global sequential models over multiple frames while also making it possible to inject monotonic development order constraints into the learning process explicitly. We demonstrate our algorithm on two time-lapse embryo video datasets: i) mouse and ii) human embryo datasets. Our method achieves 98.1\% and 80.6\% for mouse and human embryo stage classification, respectively. Our approach will enable more profound clinical and biological studies and suggests a new direction for developmental stage classification by utilizing temporal information.

\end{abstract}

\section{Introduction}
Biological developments often follow a monotonic order. A mammalian embryo's developmental process is a typical example of the monotonic constraint, which develops through cell cleavages (from 1 cell to multiple cells), morula, and blastocyst.
This monotonic constraint does not allow transitions to previous developmental stages, \textit{e.g.}, a transition from 2 cells to 1 cell.
Automated developmental stage classification can advance studying an embryo's cellular function, a basic but hard biological problem. Besides, developmental stage classification of embryos is important for \textit{in vitro} fertilization (IVF). To achieve a pregnancy, clinicians select embryos with the highest viability and transfer them to a patient. Division timing is one of the main biomarkers to assess an embryo's viability~\cite{leahy2020Automated}. 
The current standard of choosing the most promising embryos is a manual examination by clinicians via a microscope. However, manual inspection is time-consuming and prone to inter-person variability.
As such, it is essential to develop a model for automated developmental stage classification.
\begin{figure}[t]
     \centering
     \includegraphics[width=0.8\textwidth]{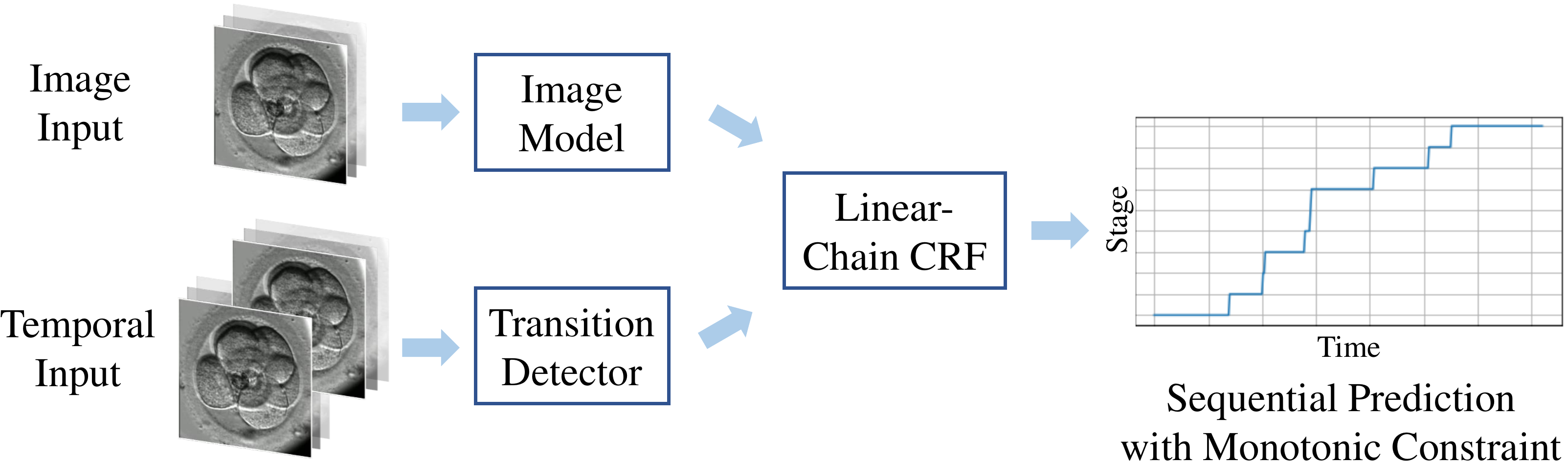}
     \caption{\textbf{Developmental Stage Classification of Embryo Time-Lapse Videos.} Our two-stream model accepts the current and the previous frames as the input. We feed the current frame into the image model. For the transition detector, we input the concatenation of the current and the previous frames to capture motion information between them. We apply the two-stream model to all the frames in a video and obtain sequential predictions using a linear-chain CRF.}
     \label{fig:teaser}
\end{figure}

In automated developmental stage classification for time-lapse videos, difficulties mainly come from overlaps between cells and imbalance between stages. Even though cells are transparent, their overlaps confuse a classifier when identifying their developmental stage. Also, a few developmental stages (\textit{e.g.,} 1, 2 cells) dominate most of the frames in time-lapse videos, which can induce class imbalance in learning.
Temporal information is valuable for addressing these two challenges. It can differentiate overlapping cells based on their movements and transitions between stages regardless of their frequencies.
Existing developmental stage classification methods~\cite{leahy2020Automated,khan2016Deep,ng2018Predicting} usually classify per-frame stages and apply dynamic programming to make use of the monotonic constraints. However, they do not incorporate temporal information, potentially solving the overlap and imbalance problems. Besides, they do not include dynamic programming in the learning process, making classification models may not learn to maximize the accuracy of dynamic programming.

In this work, we propose a two-stream model for the developmental stage classification of embryos as displayed in Fig.~\ref{fig:teaser}. We first introduce a two-stream convolutional neural network (CNN) model, which consists of an image model and a transition detector. While the image model identifies a stage of the current frame, the transition detector returns a high value when the current frame has a different label compared to the previous frame. Unlike the previous methods, we exploit temporal information in our transition detector, which can better suppress the overlap and stage imbalance issues. 
We build a linear-chain conditional random field (CRF)~\cite{sutton2012Introduction} upon our two-stream model for the monotonic constraints. Unlike conventional methods, our method effectively combines two-stream outputs using linear-chain CRF and enables learning of sequential predictions while constraining the monotonic order.
We demonstrate our algorithm's efficacy by comparing it with existing stage classification approaches on two time-lapse video datasets: i) mouse and ii) human embryos.

We have two main contributions. First, our method improves the performance for rare cell stages by combining image and temporal information in a two-stream model. Second, we inject the monotonic constraint into the learning process using linear-chain CRF to optimize the sequential predictions. 
Our code will be publicly available upon acceptance.

\section{Related Work}

\bfsection{Developmental Stage Classification of Embryos}
Researchers have proposed many stage classification methods due to their importance for IVF.
With the emergence of deep learning methods, most state-of-the-art methods rely on CNN. Khan~\etal~\cite{khan2016Deep} adopt CNN for human embryonic cell counting over the first five cell stages. Ng~\etal~\cite{ng2018Predicting} introduce late fusion nets, where multiple images are input for CNN, and additionally exploit dynamic programming to ensure a monotonical progression over time. 
Lau~\etal~\cite{lau2019Embryo} detects a region of interest and uses LSTM~\cite{gers1999Learning} for sequential classification. Rad~\etal~\cite{rad2019Cellnet} use CNN to parse centroids of each cell from embryo images. Recently, Leahy~\etal~\cite{leahy2020Automated} develop computer vision models that extract five key morphological features from time-lapse videos, including stage classification. They improve a baseline by using multiple focuses, a soft loss, and dynamic programming.

However, most previous methods focus on improving a per-frame prediction and utilize dynamic programming during testing to incorporate the monotonic development order constraint. In this work, we make use of temporal information and directly inject the monotonic condition into the learning process with CRFs for sequential stage prediction.

\bfsection{Two-Stream Models}
Researchers widely use two-stream models for action recognition. 
Two-stream 2D CNN~\cite{simonyan2014TwoStream} classifies an action by averaging predictions from image and motion branches.
3D-fused two-stream~\cite{feichtenhofer2016Convolutional} blends features from image and motion information using 3D convolution.
I3D~\cite{carreira2017Quo} replaces 2D convolutions in the two-stream 2D CNN to 3D convolutions to incorporate temporal information better. 

In their two-stream models, image and motion branches' objectives are the same; predicting an action from the input video. However, the embryo's temporal information could be useful for detecting stage transition timing rather than stage classification.
Besides, their architectural designs are for action recognition, which outputs a per-video prediction. Since embryo stage classification requires per-frame classification, the previous two-stream models may not fit sequential prediction. For sequential prediction, one may use recurrent neural networks, \textit{e.g.}, long short-term memory~\cite{gers1999Learning}. However, it is hard to incorporate the monotonic constraint of embryo development. Instead, we adopt a linear-chain CRF~\cite{sutton2012Introduction} to encode the constraints.






\begin{figure}[t]
    \centering
    \includegraphics[width=1.00\textwidth]{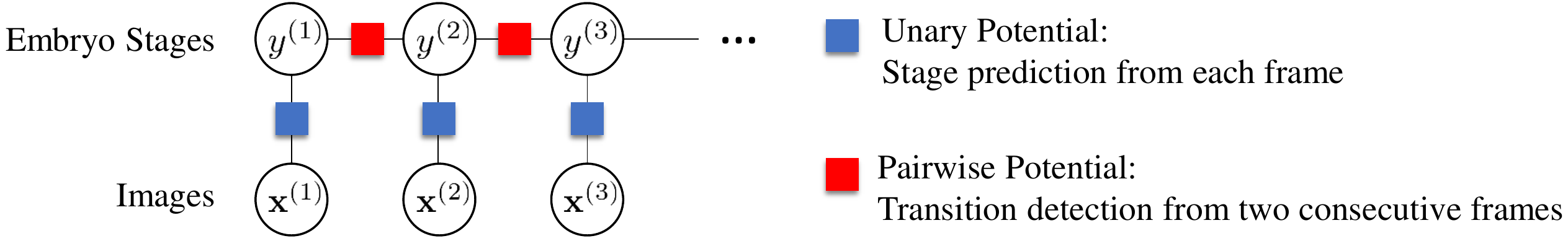}
    \caption{\textbf{Linear-Chain CRF Model.} For each image, we compute the unary potential using a image model. For pairwise ones, we use predictions from a transition detector.}
    \label{fig:method}
\end{figure}
\section{Model}
We construct a two-stream approach for the developmental stage classification of embryos. The input is a sequence of frames $X = [\textbf{x}^{(1)},\ldots,\textbf{x}^{(T)}]$, and the output is a sequence of stage predictions $Y = [y^{(1)},\ldots,y^{(T)}]$. 
As depicted in Fig.~\ref{fig:method}, we use features extracted from the two-stream model as input to a linear-chain conditional random field, where the unary potentials are from the image stream,  and the pairwise potentials are from the temporal stream. Our parameterization of the pairwise potentials (to be explained below) makes it possible to incorporate the monotonic constraint into the learning process. The entire model is trained end-to-end.

\subsection{Two-Stream Feature Encoding}
Our model uses temporal information in addition to image data to address the problems of overlapping cells and imbalance between stages, in contrast to many prior works, which often only use image information~\cite{khan2016Deep,lau2019Embryo,leahy2020Automated}. While the temporal information may not be valid for stage classification, it can be useful when there is a stage transition between two frames.
To make use of this, we adopt a two-stream approach, which consists of an \textit{image model} and a \textit{transition detector}. 
While the image model outputs scores (\textit{i.e.,} unary potentials) for each frame's stage, the transition detector outputs transition scores (\textit{i.e.,} pairwise potentials) that recognize the existence of a stage transition between two consecutive frames.

The image model infers a stage from an input frame using ResNet50~\cite{he2016Deep} pretrained on the ImageNet dataset~\cite{deng2009imagenet}. Concretely, the unary potential for class $c \in \{1, \dots, C\}$ (\textit{i.e.,} there are  $C$ possible stages) at time step $t$ is given by,
\begin{align*}
    \Phi_\textrm{I}(\mathbf{x}^{(t)} ; \theta_\textrm{I})_c = \textsf{softmax}\left(\mathbf{W}_\textrm{I} \, \textsf{ResNet}(\textbf{x}^{(t)}) + \mathbf{b}_\textrm{I} \right)_c,
\end{align*}
where $\mathbf{W}_\textrm{I}, \mathbf{b}_\textrm{I}$ are the parameters of the linear layer that outputs class scores from ResNet features, and the $\textsf{softmax}(\cdot)$ function normalizes the output to turn them into probabilities.\setcounter{footnote}{0}\footnote{Since the potentials in a CRF do not need to be probabilities, normalization via the softmax function is not strictly necessary. However, we found the normalization to be helpful for stable training. Note that if the unary potential is defined to be the output of a \emph{log}-softmax function (which is not the case in our approach), the model will reduce to a Maximum Entropy Markov Model.}

Our transition detector outputs a score for whether the current frame is in a different stage compared to the previous frame. 
Even though many two-stream methods~\cite{feichtenhofer2016Convolutional,simonyan2014TwoStream,carreira2017Quo} exploit optical flow~\cite{horn1981Determining} as temporal information, it cannot distinguish stage transition from cell movements. Hence, we feed two consecutive frames into the detector instead.
For the transition detector, we use ResNet50~\cite{he2016Deep} also pretrained on the ImageNet dataset~\cite{deng2009imagenet}, but we modify the first convolution layer to make it accept two consecutive frames as the input, $\textbf{x}^{(t-1)}$ and $\textbf{x}^{(t)}$. The detector returns a probability of stage change existence defined as,
\begin{equation*}
\rho_\textrm{M}(\textbf{x}^{(t-1)}, \textbf{x}^{(t)} ; \theta_\textrm{M})_k = 
 \textsf{softmax}\left(\mathbf{W}_\textrm{M} \, \textsf{ResNet}(\textbf{x}^{(t-1)}, \textbf{x}^{(t)}) + \mathbf{b}_{\textrm{M}}\right)_k,
\end{equation*}
where $k \in \{0, 1\}$ indicates whether there was a stage change between $\mathbf{x}^{(t-1)}$ and $\mathbf{x}^{(t)}$. The detector also implicitly parameterizes the pairwise potentials via, 
\begin{equation*}
    \Phi_\textrm{M}(\textbf{x}^{(t-1)}, \textbf{x}^{(t)} ; \theta_\textrm{M})_{(c,c')} = \begin{cases}
    \rho_\textrm{M}(\textbf{x}^{(t-1)}, \textbf{x}^{(t)} ; \theta_\textrm{M})_0, \,\, \text{if $c = c'$}\\
    \rho_\textrm{M}(\textbf{x}^{(t-1)}, \textbf{x}^{(t)} ; \theta_\textrm{M})_1, \,\, \text{if $c < c'$}\\
    -\infty, \,\, \text{otherwise},
\end{cases}     
\end{equation*}
where we penalize inverse transitions with $-\infty$ to incorporate the monotonic constraint. We use these potentials as input to the linear-chain CRF, which enables sequential classification of the input sequences taking into account the pairwise correlations that exist among the output labels.

\subsection{Linear-Chain Conditional Random Field}

We define a probability distribution over the output sequence $Y$ given the input sequence $X$ with a linear-chain CRF
\begin{equation*}
   p(Y | X; \theta_\textrm{I}, \theta_\textrm{M}) = \frac{1}{Z(X)}\prod_{t=1}^T \exp\left\{ \Phi(y^{(t-1)}, y^{(t)}, \textbf{x}^{(t)} ; \theta_\textrm{I}, \theta_\textrm{M}) \right\}, 
\end{equation*}
where $\Phi$ is a score for transitioning from $y^{(t-1)}$  to $y^{(t)}$, which is given by combining the unary and pairwise potentials from above, 
\begin{equation*}
    \Phi(y^{(t-1)}, y^{(t)}, \textbf{x}^{(t)} ; \theta_\textrm{I}, \theta_\textrm{M}) = \Phi_\textrm{I}(\textbf{x}^{(t)} ; \theta_\textrm{I})_{y^{(t)}} + \Phi_\textrm{M}(\textbf{x}^{(t-1)}, \textbf{x}^{(t)} ; \theta_\textrm{M})_{(y^{(t-1)}, y^{(t)})}.
\end{equation*}
 
Here $Z(X)$ is a normalizing constant, 
\begin{equation*}
    Z(X) = \sum_{y^{(1)}=1}^{C} ... \sum_{y^{(T)}=1}^{C} \prod_{t=1}^T \exp\left\{ \Phi(y^{(t-1)}, y^{(t)}, \textbf{x}^{(t)} ; \theta_\textrm{I}, \theta_\textrm{M}) \right\},
\end{equation*}
which can be calculated in $O(TC^2)$ with dynamic programming. 

\bfsection{Training}
During training, we also found it helpful to minimize the CRF negative log likelihood along with single-model losses derive from the image and transition models. The single-model loss for the image model is defined as,
\begin{equation*}
    \mathcal{L}_\textrm{I} = \sum_{c=1}^{C}-q^{(t)}_c\log \Phi_\textrm{I}(\textbf{x}^{(t)} ; \theta_\textrm{I})_{c},
\end{equation*}
where $\mathbf{q}^{(t)}$ is the one-hot representation of the ground truth stage at frame $t$, and the single-model loss for the transition detector is defined as,
\begin{equation*}
    \mathcal{L}_\textrm{M} = \sum_{k=0}^{1}-(1-{\textbf{q}^{(t-1)}}^T\textbf{q}^{(t)}) \log \rho_\textrm{M}(\textbf{x}^{(t-1)}, \textbf{x}^{(t)} ; \theta_\textrm{M})_c.
\end{equation*}
Thus the final loss is given by,
\begin{align*}
 -\log    p(Y | X; \theta_\textrm{I}, \theta_\textrm{M}) +  \mathcal{L}_\textrm{I} +  \mathcal{L}_\textrm{M},
\end{align*}
and we perform end-to-end training with gradient-based optimization using the Torch-struct library~\cite{rush2020Torchstruct}.\footnote{
The single-model losses and the CRF negative log likelihood are complementary each other by taking into account local and global predictions, respectively.}
We use a batch size of four and a learning rate of 0.0001 with the Adam optimizer. To construct a batch, we randomly sample 50 frames from each video and then sort them in a consecutive order. We also perform data augmentation by random resized cropping, rotation, and flipping.


\bfsection{Inference}
For prediction, our aim is in obtaining the most likely sequence of labels given a new test video $X$, \textit{i.e.},
\begin{equation*}
    \hat{Y} = \argmax_{Y} p(Y | X ; \theta_\textrm{I}, \theta_\textrm{M}).
\end{equation*}
We obtain this maximum a posteriori sequence with standard dynamic programming (\textit{i.e.,} the Viterbi algorithm). At inference time only, we also smooth the  unary potentials from the image model by modifying the potential for class $c$,
\begin{align*}
\frac{1}{13}\cdot \Phi_\textrm{I}[c-2] + \frac{3}{13}\cdot \Phi_\textrm{I}[c-1] +    \frac{5}{13} \cdot \Phi_\textrm{I}[c] +
\frac{3}{13} \cdot \Phi_\textrm{I}[c+1] +
\frac{1}{13} \cdot \Phi_\textrm{I}[c+2],
\end{align*}
and using the above weighted average as the input to the Viterbi algorithm.
This reweighting values, which were found via a search on the validation set, take into account the ordinal nature of the output space (boundaries are zero-padded). 

\section{Experimental Results}
We evaluate our method's performance, demonstrating each design choice's effect in the models with ablation studies. 
We evaluate stage classification algorithms on two embryo datasets: i) mouse and ii) human embryo datasets. 

\bfsection{Compared Methods}
We compare our method with a general classification model, ResNet50~\cite{he2016Deep}, one state-of-the-art embryo stage classification method, AutoIVF~\cite{leahy2020Automated}, an early fusion method~\cite{karpathy2014Largescale} that leverages temporal information, and a sequential model~\cite{lau2019Embryo} based on LSTM. For a fair comparison, we re-implement AutoIVF using a single focus and the same backbone as ours. The early fusion takes five successive frames as input and learns to predict the middle frame's stage. We adopt the PyTorch 1.7 library to implement all the methods.

\bfsection{Evaluation Metric}
We evaluate classification accuracy as the number of correct predictions over the number of data (Global). Since the majority stages, such as 1 cell and 2 cells, can dominate the average accuracy, we calculate the per-stage accuracies and the mean of them (Per-Stage). We train all methods for five seeds and report their average performances with standard deviations.

\subsection{Developmental Stage Classification of Mouse Embryos}

\bfsection{Dataset} We use the NYU Mouse Embryo dataset consisting of 100 videos of developing mouse embryos~\cite{cicconet2014Labela}. The videos contain 480 x 480 resolution images taken every seven seconds, with a median of 314 frames per embryo, totaling an average length of 36.6 minutes per embryo. 
The videos have frames with up to 8 cells, \textit{i.e.,} eight developmental stages. 
For training and evaluation, we randomly split the data 80/10/10 into train, validation, and test videos, respectively. We use the validation set to select hyper-parameters and models for evaluation. 

\bfsection{Result}
In Table~\ref{table:mouse}, we list overall and per-stage classification performances of the embryo stage classification methods. Our method outperforms all other methods on average for various stages. The frequency imbalance between the stages allows LSTM to achieve comparable results on average over all the data. 


\bgroup
\def\arraystretch{1}
\begin{table}[t]
    \centering
    \caption{Accuracies (\%) of stage classification methods on the test mouse embryos~\cite{cicconet2014Labela}.}
    \resizebox{0.83\textwidth}{!}{
    \begin{tabular}{lcccccccccc}
    \hline
    Method & Global & Per-Stage & 1 & 2 & 3 & 4 & 5 & 6 & 7 & 8\\
    \hline
    \multicolumn{11}{l}{\textcolor{mygray}{A. Per-Image Classification Model}}\\
    \hline
    ResNet50~\cite{he2016Deep} & 90.9$\pm$0.6 & 59.4$\pm$1.2 & 99.0 & 98.3 & 89.1 & 90.7 & 25.1 & 15.8 & 26.5 & 30.4 \\
    AutoIVF~\cite{leahy2020Automated} & 96.4$\pm$0.1 & 60.9$\pm$3.1 & 99.8 & \bf{99.9} & 92.3 & \bf{99.9} & 4.3 & 33.5 & 50.8 & 6.2\\
    \hline
    \multicolumn{11}{l}{\textcolor{mygray}{B. Spatiotemporal Classification Model}}\\
    \hline
    Early Fusion~\cite{karpathy2014Largescale} & 91.9$\pm${0.1} & 57.1$\pm$0.2 & 98.8 & \bf{99.9} & 74.0 & 93.2 & 0.0 & 14.9 & 37.9 & \bf{37.6}\\
    LSTM~\cite{lau2019Embryo} & 98.0$\pm$0.1& 45.0$\pm$1.7 & 96.9 & 98.7 & 22.9 & 42.2 & 0.0 & 9.2 & \bf{90.1} & 0.0 \\
    Ours & \bf{98.1$\pm$0.3} & \bf {76.8$\pm$5.4} & \bf{99.9} & \bf{99.9} & \bf{94.9} & \bf{99.9} & \bf{35.1} & \bf{84.6} & 71.4 & 29 \\
    \hline
    \end{tabular}}
\label{table:mouse}
\end{table}
\egroup






\subsection{Developmental Stage Classification of Human Embryos}

\bfsection{Dataset}
We evaluate the stage classification methods on the human embryo dataset~\cite{leahy2020Automated}. There are 13 stage labels: empty well, 1 cell to 9+ cells, morula (M), blastocyst (B), and degenerate embryo. To focus on the embryo development's monotonicity, we only use 11 stages, excluding frames with the empty well and degenerate embryo labels.
The dataset includes 341 training, 73 validation, and 73 test time-lapse videos of embryos. Each video consists of 325 frames on average. As the network input, we crop zona-centered patches from each frame to exclude outside regions of interest and resize the frames to 112 $\times$ 112 resolution.

\bfsection{Result}
Table~\ref{table:human} benchmarks the developmental stage classification methods. Overall, our approach surpasses the other classification methods. In terms of the mean per-stage accuracy, the performance gain over the existing methods is much higher, which indicates our method notably performs better for rare developmental stages. 
Since we incorporate the transition detector and use it to force the predictions of our model to be monotonic, our method outperforms the two spatiotemporal methods; Early Fusion and LSTM. Unlike AutoIVF, our model learns the features for the stage change detection, which are helpful for the monotonic predictions.

\begin{figure}[t]
    \centering
    \subfigure[AutoIVF: 86.6\% / Ours: 95.3\%]{\includegraphics[width=0.49\textwidth]{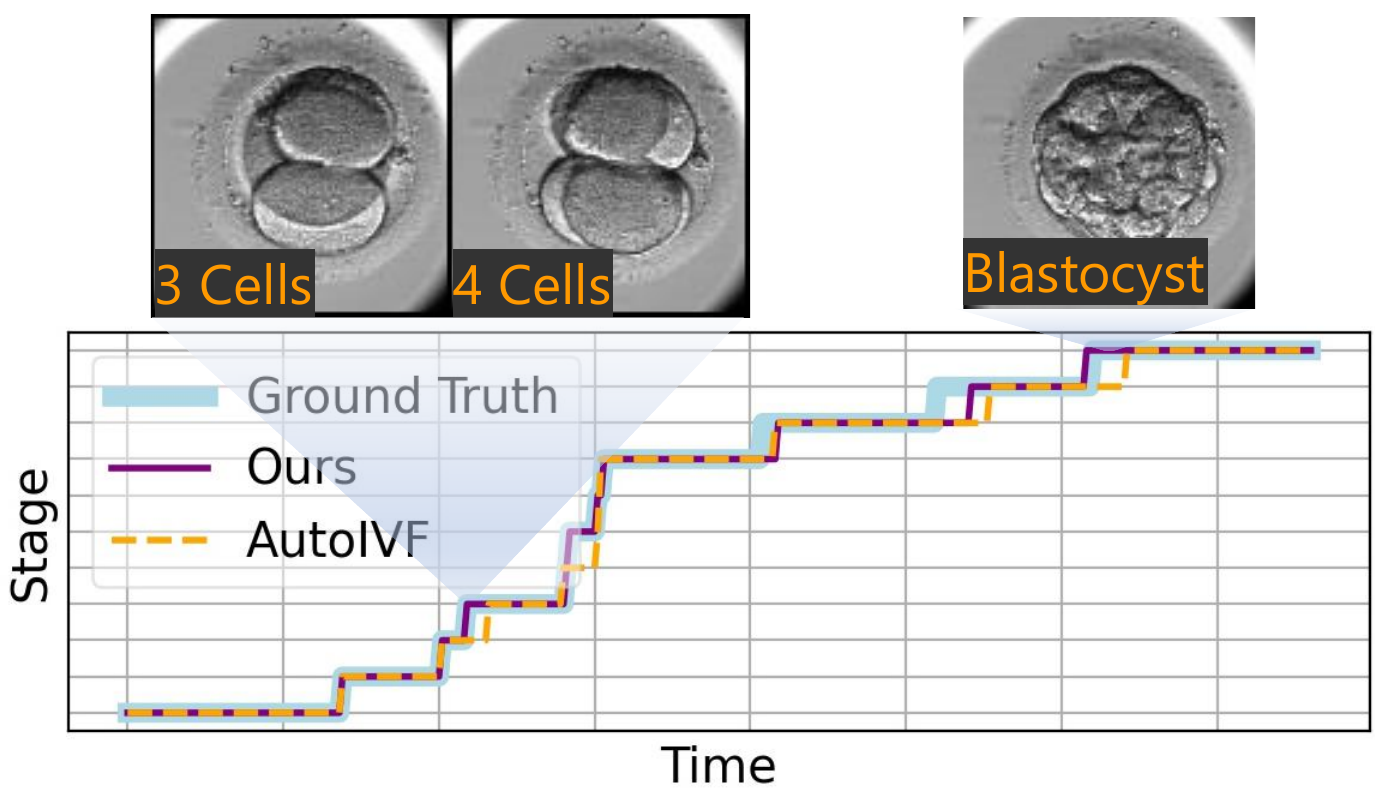}}
    \subfigure[AutoIVF: 85\% / Ours: 92.9\% ]{\includegraphics[width=0.49\textwidth]{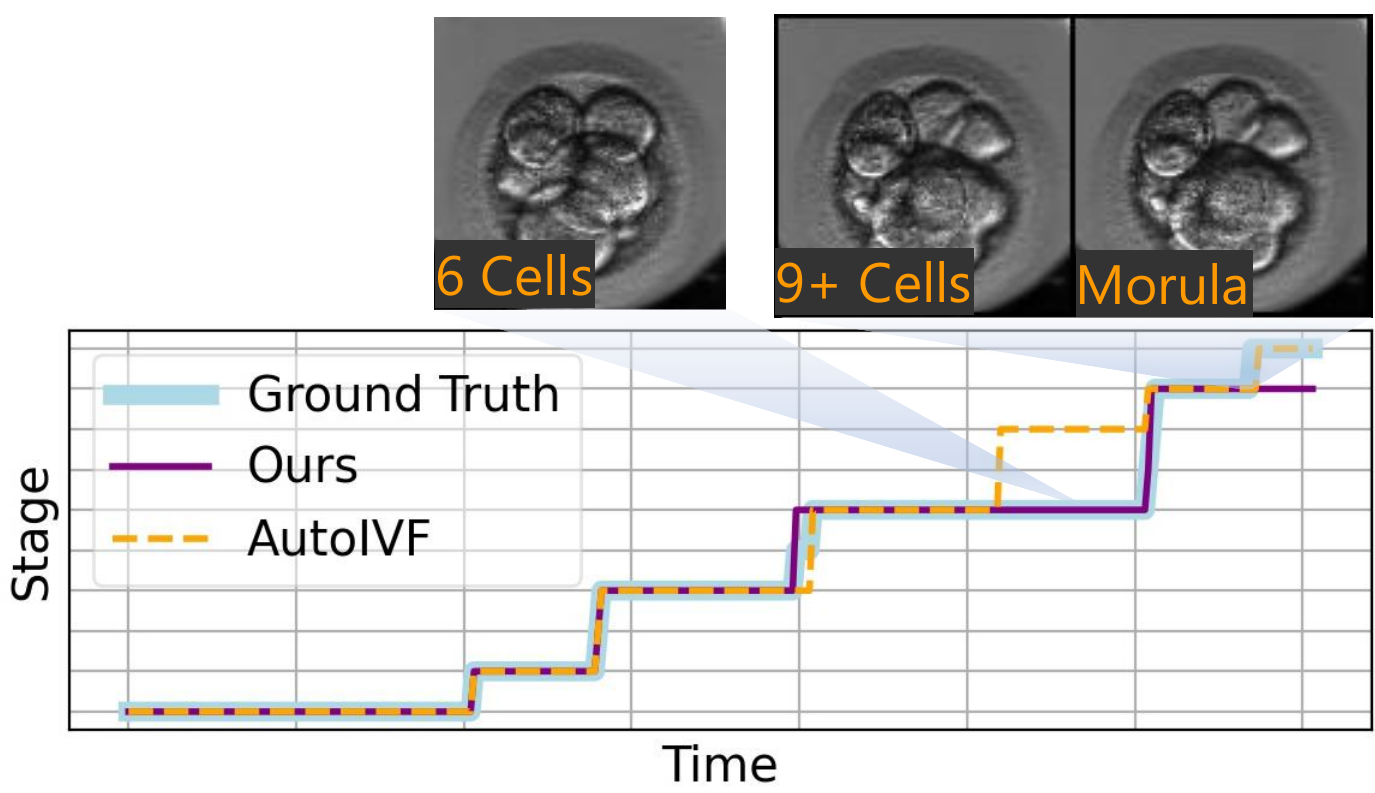}}
    \caption{\textbf{Qualitative Stage Classification Results.} We visualize frames with ground truths (lower left corner), where our method and AutoIVF~\cite{leahy2020Automated} predict different stages.} 
    \label{fig:qualitative}
\end{figure}

Our method runs in 268 frames per second on a single TITAN X GPU. Our model has 47M parameters and requires up to 4 GB GPU memory in the inference phase. 
Fig.~\ref{fig:qualitative} visually compares our method with AutoIVF~\cite{leahy2020Automated}. Our method is better at detecting cell division timings. As one example of failure cases, our model fails to detect the transition between 9+ cells and morula in Fig.~\ref{fig:qualitative} (b) since it takes two consecutive frames as the input, which visually have no major difference in this example.

\bgroup
\def\arraystretch{1}
\begin{table}[t]
    \centering
    \caption{Scores (\%) of stage classification methods on the test human embryos~\cite{leahy2020Automated}.}
    \resizebox{1.0\textwidth}{!}{
    \begin{tabular}{lccccccccccccc}
    \hline
    Method & Global & Per-Stage & 1 & 2 & 3 & 4 & 5 & 6 & 7 & 8 & 9+ & M & B\\
    \hline
    \multicolumn{14}{l}{\textcolor{mygray}{A. Per-Image Classification Model}}\\
    \hline
    ResNet50~\cite{he2016Deep} & 74.6$\pm$1.0 & 58.2$\pm$1.6 & 97.6 & 93.8 & 24.3 & 80.8 & 24.1 & 16.2 & 19.8 & 55.2 & 63.5 & 70.7 & 93.9 \\
    AutoIVF~\cite{leahy2020Automated} & 77.8$\pm$1.2 & 60.9$\pm$2.2 & 98.2 & \textbf{96.6} & 22.9 & 88.2 & 26.5 & 15.6 & 22.4 & 59.3 & 67.3 & 77.0 & 96.1 \\
    \hline
    \multicolumn{14}{l}{\textcolor{mygray}{B. Spatiotemporal Classification Model}}\\
    \hline
    Early Fusion~\cite{karpathy2014Largescale} & 75.1$\pm0.6$ & 55.7$\pm0.7$ & 97.5 & 93.4 & 10.2 & 84.5 & 11.5 & 7.9 & 12.8 & 63.5 & 65.7 & 72.5 & 93.7\\
    LSTM~\cite{lau2019Embryo} & 77.1$\pm$0.9 & 61.8$\pm$0.9 & 97.8 & 92.7 & 31.4 & 79.9 & 21.4 & 25.4 & \textbf{28.8} & 58.3 & \textbf{67.6} & \textbf{79.3} & \textbf{97.0}\\
    Ours & \textbf{80.6$\pm$0.7} & \textbf{66.3$\pm$1.9} & \textbf{99.4} & 96.2 & \textbf{41.2} & \textbf{89.4} & \textbf{43.3} & \textbf{27.6} & 19.7 & \textbf{69.8} & 67.0 & 78.7 & 96.7 \\
    \hline
    \end{tabular}}
\label{table:human}
\end{table}
\egroup


\subsection{Ablation Study}

We analyze our method's efficacy by conducting an ablation study on the human embryo dataset. To this end, we add one of our components to the baseline at a time. By performing dynamic programming without pairwise potentials, our model improves the baseline's accuracy from 76.7\% to 80.0\%. Using both unary and pairwise terms in linear-chain CRF, our two-stream model yields 80.6\% score, which performs the best. In conclusion, our full setting enables the maximum performance for developmental stage classification.




\section{Conclusion and Future Work}
Our method will enable better clinical and embryological studies by improving the accuracies on rare stages, which are infrequent in videos but equally important as frequent stages when analyzing embryos. 
Since we measure stage transition probabilities, cell division timings predicted by our method are highly interpretable, which will allow tractable inspection in clinical practice.
Our future work includes further improving performance on rare stages by combining a stage classifier and a cell detector, developing sequential models for other developmental features of embryos, and experimenting with different ways of acquiring unary and pairwise potentials, \textit{e.g.,} calculating the transition probability over longer sliding windows of frames.



\section*{Acknowledgements}
This work was funded in part by NIH grants 5U54CA225088 and R01HD104969, NSF Grant NCS-FO 1835231, the NSF-Simons Center for Mathematical and Statistical Analysis of Biology at Harvard (award number 1764269), the Harvard Quantitative Biology Initiative, and Sagol fund for studying embryos and stem cells; Perelson Fund.

\bibliographystyle{splncs04}
\bibliography{MICCAI2021-StageClassification}

\end{document}